\documentclass[final,3p,times,twocolumn]{elsarticle}
\usepackage{amssymb}
\usepackage{lineno}
\usepackage{amsmath,amsfonts}
\usepackage{algorithmic}
\usepackage{algorithm}
\usepackage{array}
\newcolumntype{P}[1]{>{\centering\arraybackslash}p{#1}}
\usepackage[caption=false,font=normalsize,labelfont=sf,textfont=sf]{subfig}
\usepackage{textcomp}
\usepackage{stfloats}
\usepackage{url}
\usepackage{verbatim}
\usepackage{graphicx}
\usepackage{hyperref}
\usepackage{epsfig}
\usepackage{graphicx}
\usepackage{amsmath}
\usepackage{amssymb}
\usepackage{booktabs}
\usepackage{stfloats}
\usepackage{float}
\usepackage{pifont}
\usepackage{enumitem}
\usepackage{multirow}
\usepackage{bm}
\usepackage{bbding}
\usepackage{etoolbox}
\usepackage{wrapfig}
\usepackage{graphicx}
\usepackage{subcaption}
\usepackage{color, colortbl}
\usepackage{xcolor}
\usepackage[dvipsnames]{xcolor}
\definecolor{Gray}{gray}{0.9}

\definecolor{mygreen}{RGB}{114, 210, 126}

\usepackage[capitalize]{cleveref}
\crefname{section}{Sec.}{Secs.}
\Crefname{section}{Section}{Sections}
\Crefname{table}{Table}{Tables}
\crefname{table}{Tab.}{Tabs.}

\journal{Pattern Recognition}

\begin{document}

\begin{frontmatter}

\title{UltraSeP: Sequence-aware Pre-training for Echocardiography Probe Movement Guidance}

\author[thu,baai]{Haojun Jiang}\corref{cor1}
\author[thu]{Teng Wang}\corref{cor1}
\author[baai]{Zhenguo Sun}
\author[thu]{Yulin Wang}
\author[thu]{Yang Yue}
\author[baai]{Yu Sun}
\author[baai]{Ning Jia}
\author[baai]{Meng Li}
\author[baai]{Shaqi Luo}
\author[thu]{Shiji Song}
\author[thu]{Gao Huang \corref{cor2}}

\affiliation[thu]{organization={Department of Automation, BNRist, Tsinghua University},
            postcode={100084}, 
            state={Beijing},
            country={China}}

\affiliation[baai]{organization={Beijing Academy of Artifical Intelligence},
            postcode={100084}, 
            state={Beijing},
            country={China}}

\cortext[cor1]{This work was done while Haojun Jiang was an intern at Beijing Academy of Artificial Intelligence.}
\cortext[cor2]{Corresponding Author}
\ead{jhj20@mails.tsinghua.edu.cn or jianghaojunthu@163.com (Haojun Jiang); gaohuang@tsinghua.edu.cn}

\begin{abstract}
Echocardiography is an essential medical technique for diagnosing cardiovascular diseases, but its high operational complexity has led to a shortage of trained professionals.
To address this issue, we introduce UltraSeP, a novel probe movement guidance algorithm that has the potential to be applied in guiding robotic systems or novices with probe pose adjustment for high-quality standard plane image acquisition.
Cardiac ultrasound faces two major challenges: (1) the inherently complex structure of the heart, and (2) significant individual variations.
Previous works have only learned the population-averaged structure of the heart rather than personalized cardiac structures, leading to a performance bottleneck.
Clinically, we observe that sonographers dynamically adjust their interpretation of a patient's cardiac anatomy based on prior scanning sequences, consequently refining their scanning strategies.
Inspired by this, we propose a novel sequence-aware self-supervised pre-training method. 
Specifically, our approach learns personalized three-dimensional cardiac structural features by predicting the masked-out image features and probe movement actions in a scanning sequence.
We hypothesize that if the model can predict the missing content it has acquired a good understanding of personalized cardiac structure.
Extensive experiments on a large-scale expert scanning dataset with 1.67 million samples demonstrate that our proposed sequence-aware paradigm can effectively reduce probe guidance errors compared to other advanced baseline methods.
\end{abstract}

\end{frontmatter}

\section{Introduction}
\label{sec:introduction}
Cardiovascular disease is the leading threat to human health \cite{roth2017global}.
The most common method for assessing cardiac health is transthoracic echocardiography, as recommended by the American Society of Echocardiography~\cite{mitchell2019guidelines}.
In this procedure, a sonographer manually maneuvers a probe to find critical two-dimensional planes and observes the myocardium, valves, and blood flow~\cite{braunwald2015war} in these planes (views) in real-time using ultrasound images.
However, due to the complexity of the heart's structure, even small changes in the 6 degrees of freedom (6-DOF) pose of the probe can affect the imaging of the structures being observed. 
Furthermore, individual differences in heart size, shape, and structure~\cite{wild2017large} make it even more challenging, requiring adjustments to scanning strategies based on each patient's unique characteristics.
This places high demands on the sonographer's scanning skills, leading to a long learning curve for sonographers and a shortage of practitioners~\cite{won2024sound}.

\begin{figure}[t!]
\centering
\includegraphics[width=1\columnwidth]{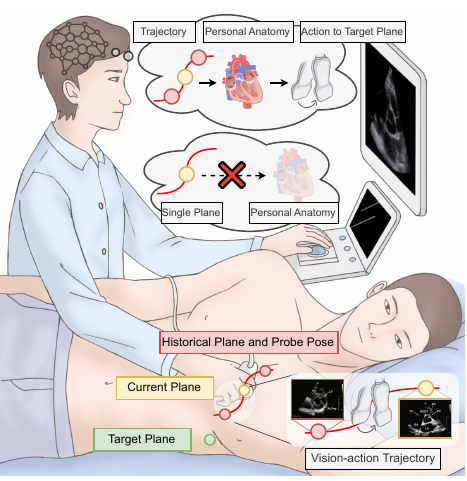}
\caption{
Motivation of our work.
During scanning, sonographers develop a cognitive understanding of the patient's cardiac anatomy using past trajectories (plane images and probe poses), which guides subsequent scanning.
}
\label{fig1:motivation}
\end{figure}

Previous researchers proposed assistive scanning systems~\cite{jiang2024cardiac,narang2021utility,bao2024real}, which provide 6-DOF pose adjustment suggestions for the probe based on the current ultrasound image to reach the target plane.
These system has the potential to improve the quality and efficiency with which inexperienced sonographers find standard planes, thereby increasing the global supply of echocardiography.
These works have attempted to model the population-averaged cardiac structure from large-scale data and rely solely on a single plane's images to make action decisions.
This paradigm neglects individual variations in cardiac structure, leading to a performance bottleneck.
Clinically, as shown in \cref{fig1:motivation}, sonographers learn personalized cardiac structural information based on the acquired planes and their spatial pose information from the patient~\cite{laumer2023weakly,chen20253d}, further assisting in subsequent scans.
From this observation, we can derive two key insights: (1) sonographers make decisions based on previous sequence data (plane images and corresponding probe pose) rather than a single image, and (2) sequence data contains personalized information about cardiac structure.
Several researchers~\cite{droste2020automatic,men2023gaze} have also explored using sequences as network inputs. 
However, we empirically find that simply increasing the sequence length does not lead to continuous performance improvement (Sec.~\ref{sec:exp}).  
Therefore, models does not inherently possess the ability to extract beneficial individual cardiac structural information from sequences.
To this end, we propose a novel sequence-aware self-supervised pre-training method, namely UltraSeP, which learns personalized cardiac structures through a masked prediction task.
To be specific, we sample a scanning trajectory comprising plane images and the probe movement actions between them, which are derived from the probe's pose, from the same patient. 
Both the plane images and actions are processed by modality-specific encoders to obtain corresponding features, with some features randomly masked as targets for prediction.
For the masked images, we apply supervision in the feature space, as \cite{assran2023self} demonstrated that this is a more efficient learning method. 
The actions, on the other hand, are re-predicted to their original values for supervision.
This paradigm requires the model to predict the post-action plane visual feature based on the given probe movement action and visible patient's cardiac structure context information, or to predict the probe movement action between two planes.
Thus, this method enables the model to acquire the ability to construct an implicit three-dimensional cardiac structure based on known patient information.

\begin{figure*}[t!]
\centering
\includegraphics[width=2.1\columnwidth]{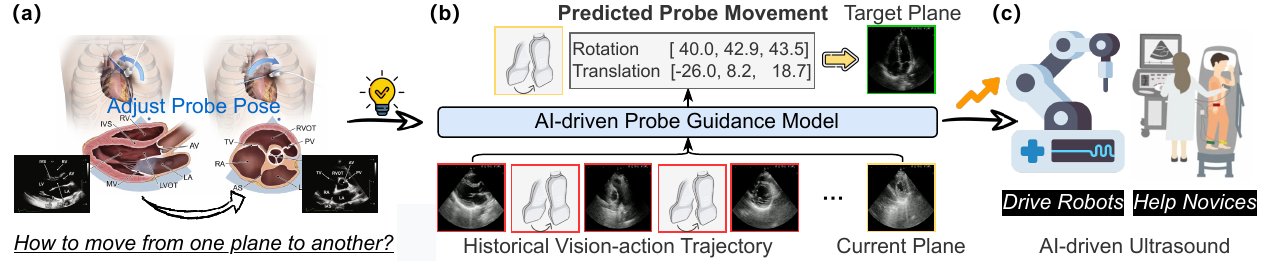}
\caption{
Illustration of the probe movement guidance task.
(a) Cardiac ultrasound requires probe pose adjustment to capture different standard plane images.
(b) We develop a guidance model that predicts probe movements for target plane acquisition from scan trajectories.
(c) With the guidance of this model, ultrasound robotic system or novices have the potential to perform cardiac ultrasound examinations more effectively.
}
\label{fig2:task_illustration}
\end{figure*}

The probe movement guidance task aims to provide sonographers with probe adjustment suggestions to accurately locate target planes, as shown in \cref{fig2:task_illustration}. 
This task demands a high level of cardiac structure modeling capability, making it an effective way to assess whether the pre-training has successfully imparted a generalizable understanding of cardiac structure.
In this downstream task, we also introduce a sequence modeling approach, which predicts the next probe movement action based on ultrasound images and actions from historical scan data.
Furthermore, loading our sequence-aware model pre-trained on a dataset containing 284 routine scanning trajectories (1.03 million samples) significantly improves the probe guidance performance.
Experiments on a validation set containing 72 routine scanning trajectories (0.28 million samples) demonstrate that our method outperforms baseline approaches. 
Besides, detailed ablation studies validate the effectiveness of critical design component and the impact of various hyper-parameters on our method.
In summary, this paper makes three-fold contributions:
\begin{itemize}[itemsep=0pt]
    \item We introduce a cardiac probe movement guidance model (UltraSeP) that aims to assist in acquiring high-quality standard plane images. To the best of our knowledge, this is the first guidance model that is technically disclosed and capable of covering up to ten critical standard planes.
    \item We innovatively propose a sequence-aware self-supervised pre-training method that enhances the model's ability to capture personalized cardiac structure information from historical scanning sequences.
    \item Extensive experiments conducted on a large-scale dataset containing 1.67 million clinical expert-scanned samples demonstrate that our method consistently outperforms baseline approaches across various experimental settings, providing more accurate guidance signals.
\end{itemize}

\section{Related Work}
With the advancement of AI technologies~\cite{dosovitskiy2020image,he2022masked,huang2017densely}, AI-assisted applications have shown significant potential in echocardiography, such as ejection fraction estimation~\cite{ouyang2020video}, and anatomical structure segmentation~\cite{zhang2023slide,ding2024c2fresmorph}

\textbf{AI-assisted ultrasound scanning.}
To achieve AI-assisted scanning, one line of research adopts reinforcement learning methods~\cite{amadou2024goal,li2023rl,bi2022vesnet}, while another focuses on the supervised learning paradigm~\cite{narang2021utility,droste2020automatic,jiang2025towards}.

For reinforcement learning methods, the primary challenge lies in constructing a good simulation environment.
Current approaches~\cite{amadou2024goal,li2023rl,bi2022vesnet} mainly use computed tomography (CT) scan data of the heart as the simulation environment.
For example, Amadou et al.~\cite{amadou2024goal} first segments anatomical structures from the CT data to identify the cardiac region, and then reconstructs this region into an ultrasound volume as training environment.
A similar strategy is adopted in works~\cite{li2023rl}.
However, acquiring cardiac CT scan data is more costly than ultrasound data, which limits the scalability of such methods.
Additionally, differences between CT and ultrasound data pose challenges for domain adaptation.
Bi et al.~\cite{bi2022vesnet} also attempted to artificially construct a simulation environment for vascular organs. However, such manually constructed simulation environments struggle to model the anatomical variations present in real clinical settings, particularly for organs like the heart, which exhibit significant inter-individual variability.
Therefore, although reinforcement learning strategies hold significant potential, their development in ultrasound-assisted scanning remains relatively limited due to the difficulty in constructing realistic simulation environments.

For supervised learning methods, expert scanning demonstration data is typically collected, where each plane image is paired with a corresponding probe pose or adjustment action.
This data is used to learn the mapping between plane images and the corresponding probe movement actions.
Recent works~\cite{jiang2025towards} has demonstrated that this strategy shows promising generalization performance in vascular scanning.
Specifically, Narang et al.~\cite{narang2021utility} proposed a single-frame based commercial AI algorithm, trained on 500K labeled samples, to guide novices in performing echocardiography.
However, this is a clinical-oriented work, and the technical details have not been fully disclosed. 
Unlike them, we propose a novel algorithm and make their details publicly available.
Bao et al.~\cite{bao2024real} proposed a navigation system that provides only probe rotation movement guidance for locating the apical four-chamber (A4C) and two-chamber  (A2C) plane.
However, considering only the A4C and A2C planes is far from sufficient to meet clinical needs. 
Besides, transitioning from the A4C plane to the parasternal long-axis plane involves a complex combination of translational and rotational motions, making it highly restrictive to focus solely on rotation.
Jiang et al.~\cite{jiang2024cardiac,jiang2024structure} proposed pre-training strategies to learn the implicit cardiac structure, which is helpful for the downstream probe guidance task.
Furthermore, Jiang et al. did not consider the issue of individual differences.
Meanwhile, our paper proposes a novel pre-training method to enhance the model's ability to construct personalized cardiac structures.
Droste et al.~\cite{droste2020automatic} proposed US-GuideNet and Men et al.~\cite{men2023gaze} proposed Multimodal-GuideNet. 
Both approaches use sequence as input to guide obstetric ultrasound scanning.
Empirically, we find that simply increasing the sequence length does not lead to performance improvements, indicating that the model’s ability to capture individual structural information from sequences needs optimization (Sec.~\ref{sec:exp}).
However, US-GuideNet and Multimodal-GuideNet do not focus on enhancing the model's ability to extract useful information from past sequences, resulting in suboptimal performance.
Moreover, they also focuses exclusively on rotation, which represents a major limitation of these works.

\textbf{AI-assisted ultrasound diagnosis.}
Unlike scanning, AI-assisted ultrasound diagnosis~\cite{christensen2024vision,jiao2024usfm,dong2026hypergraph,chen2023rethinking,peng2024multi} has developed much faster due to the abundance of available data.
For instance, Christensen et al. \cite{christensen2024vision} proposed a vision-language foundation model EchoCLIP for echocardiogram interpretation.
EchoCLIP is a strong cardiac diagnosis model trained with over 1 million video-text pairs from 224K echocardiography studies across a decade of clinical care.
Meanwhile, Jiao et al. proposed USFM~\cite{jiao2024usfm}, a universal ultrasound foundation model designed for multi-task and multi-organ applications.
These two powerful diagnostic foundation models \cite{christensen2024vision,jiao2024usfm} learn visual features useful for ultrasound scanning, so we include them in the experimental section for comparison.

\section{Dataset and method}
In this section, we first describe the collected echocardiographic scanning dataset to provide a background about our task.
Next, we introduce the proposed sequence-aware pretraining framework, designed to learn the ability to model personalized cardiac structures from historical scanning sequences.
Lastly, the pre-trained model is fine-tuned to perform the downstream probe movement guidance task.

\begin{figure*}[!t]
\centering
\includegraphics[width=2.1\columnwidth]{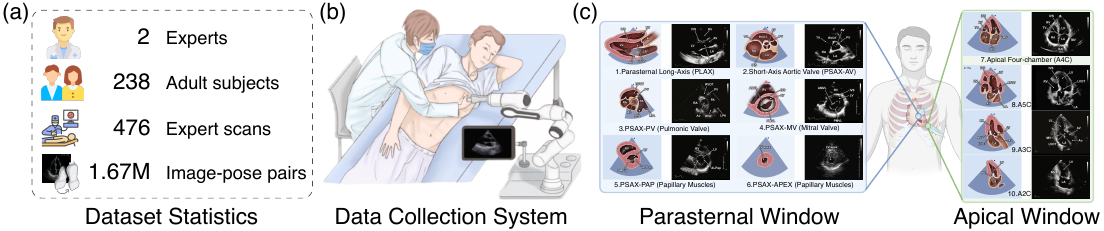}
\caption{
Echocardiographic scanning dataset.
(a)
Dataset statistics. The dataset was collected by two senior sonographers who performed scans on 238 adult subjects, two scans per individual. 
This resulted in a large-scale dataset containing 1.67 million samples.
(b)
Dataset collection system. 
The sonographer operated a probe attached to the end of a robotic arm for scanning.  
The system records the ultrasound images along with the corresponding probe poses.
(c) 
Collected ten standard planes.
The sonographer scanned six standard planes from the parasternal window and four standard planes from the apical window.
Images are sourced from the guideline~\cite{mitchell2019guidelines}.
}
\label{fig3:dataset}
\end{figure*}

\subsection{Echocardiographic Scanning Dataset}
\label{method:dataset}
To conduct the research, we collected a large-scale dataset comprising 476 expert scanning trajectories from 238 adult subjects, performed by two sonographers with over 10 years of experience.
To collect the data, we constructed a system as shown in \cref{fig3:dataset}b, where sonographers manipulated a probe attached to the end of a robotic arm to perform continuous scans of ten crucial echocardiogram planes (\cref{fig3:dataset}c, recommended by the American Society of Echocardiography~\cite{mitchell2019guidelines}).
The system simultaneously recorded the images captured by the probe and the corresponding 6 degrees of freedom (6-DOF) pose data, forming a scan sequence $\{(\mathbf{I}_t, \mathbf{p}_t)\}^{T}_{t=1}$, where $t$ is the timestamp, $\mathbf{I}_t \in\mathbb{R}^{C \times H \times W}$ is the ultrasound image at time $t$, and $\mathbf{p}_t \in\mathbb{R}^{6}$ is the corresponding probe pose.
The probe’s 6-DOF pose is represented by the translational coordinates along the $x$, $y$, and $z$ axes, and the corresponding rotational Euler angles around the $x$, $y$, and $z$ axes.
The values of translation (mm) and rotation (°) are of the same order of magnitude.
Since the probe belongs to the rigid transformation group, we can calculate the relative movement action $\mathbf{a}_{i \rightarrow j}$ between two probe poses $\mathbf{p}_{i}$ and $\mathbf{p}_{j}$.

In each scan, sonographers label the timestamp $t_g$, where $g$ ranges from 1 to 10, corresponding to ten standard planes.
Given the timestamp $t_g$, we can obtain the probe pose $\mathbf{p}_{t_g}$ corresponding to the $g$-th standard plane.
Then, the probe movement $ \mathbf{a}_{t_j \rightarrow t_g} $ from other plane $ (\mathbf{I}_{t_j}, \mathbf{p}_{t_j}),t_j \neq t_g$ to the standard plane can be calculated. 
This value further serves as the ground truth (supervision) for the probe movement guidance task.

\subsection{UltraSeP: Sequence-aware Pre-training}
The core intuition is that a patient's scanning sequence contains personalized cardiac information. 
If the model can accurately predict the masked parts, it indicates that the model has a good understanding of the patient's cardiac structure.
Below, we describe the details of our method.

\begin{figure*}[t!]
\centering
\includegraphics[width=2.1\columnwidth]{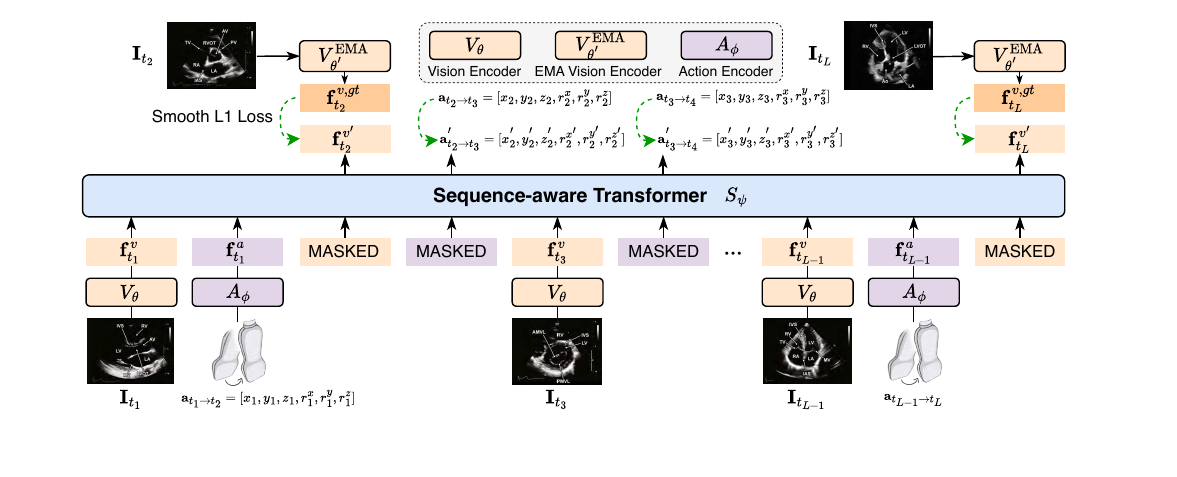}
\caption{
Diagram illustrating the sequence-aware pre-training method.
Please zoom in to view. 
The input consists of a scanning sequence, including ultrasound images and corresponding relative movement actions between them.
A portion of the input is randomly masked, and the model is required to recover masked visual features and actions.
}
\label{fig4:framework}
\end{figure*}

\textbf{Input.}
Firstly, given an image $ I_t $ from a specific scan, we sample $ L-1 $ additional images and their corresponding poses from the same scan.
The sampling strategy will be explained in \cref{method:sampling}.
The $ L-1 $ sampled images are sorted in ascending order based on their timestamps, with $ I_t $ positioned at the end of the sequence as $ I_{t_L} $. 
Then, their relative movement actions are calculated based on their poses, thus generating a scanning trajectory:
\begin{align}
    [\mathbf{I}_{t_1}, \mathbf{a}_{t_1 \rightarrow t_2}, \cdots,  \mathbf{I}_{t_{L-1}}, \mathbf{a}_{t_{L-1} \rightarrow t_{L}}, \mathbf{I}_{t_L}], 
\end{align}
where $\mathbf{I}_{t_i}\in\mathbb{R}^{C \times H \times W}, \mathbf{a}_{t_i \rightarrow t_j}\in\mathbb{R}^{6}$.
Next, we apply a random mask to the elements in the sequence, regardless of modality, with a mask ratio between 0.3 and 0.5, as shown in \cref{fig4:framework}.
In our method, the mask ratio is smaller compared to approaches like Masked Autoencoder~\cite{he2022masked} because the elements in our sequence are linked through actions and image features, rather than being constrained by the natural spatial continuity found in images.
Therefore, if the mask ratio is too high, the model loses the necessary constraints for reasoning. 
For instance, if only the first image remains unmasked, the model is unable to infer the subsequent trajectory and, consequently, cannot learn the cardiac structural knowledge contained within the scanning trajectory.

The visible image input $\mathbf{I}_{t_i}$ is first divided into $N$ non-overlapping patches $\mathbf{I}_{t_i}^{\mathrm{p}} \in \mathbb{R}^{N \times (P^2 \cdot C)}$.
Then, the $\mathbf{I}_{t_i}^{p}$ are processed by the vision transformer $V_{\theta}$~\cite{dosovitskiy2020image}:
\begin{align}
    \mathbf{H}_{t_i}^\mathrm{v} &= V_\theta(\mathbf{I}_{t_i}^{\mathrm{p}}\mathbf{E} + \mathbf{E}^\mathrm{pos}),&\mathbf{E}\in\mathbb{R}^{(P^2\cdot C) \times D}, \ \mathbf{H}_{t_i}^{\mathrm{v}},\mathbf{E}^\mathrm{pos}\in\mathbb{R}^{N \times D},
\end{align}
where $\mathbf{E}$ is implemented as a fully connected layer to convert patches to vectors and $\mathbf{E}^\mathrm{pos}$ is a learnable position embedding, and $D$ is hidden dimension.
Then, we average the patches' feature $\mathbf{H}_{t_i}^{\mathrm{v}}$ to obtain a global representation, because direct input of original image patches to the transformer would cause prohibitively long sequences, leading to substantially higher computational overhead.
Simultaneously, timestep embeddings, which represent the positional information within the sequence, are added to the global feature:
\begin{align}
    \mathbf{f}_{t_i}^{\mathrm{v}} &= \mathbf{T}_{i} + \frac{1}{N}\sum_{n=1}^{N}\mathbf{H}_{t_i,n}^{\mathrm{v}}, \ \ &\mathbf{f}_{t_i}^{\mathrm{v}}, \mathbf{T}_i \in \mathbb{R}^{D}, 
\end{align}
where $\mathbf{T} \in \mathbb{R}^{L \times D}$ is the timestep embedding.
The visible relative movement action $\mathbf{a}_{t_j \rightarrow t_{j+1}}$ is similarly processed through an action encoder $A_\phi$:
\begin{align}
    \mathbf{f}_{t_j}^{\mathrm{a}} &= \mathbf{T}_{j} + A_\phi(\mathbf{a}_{t_j \rightarrow t_{j+1}}), \ \ &\mathbf{f}_{t_j}^{\mathrm{a}}, \mathbf{T}_j \in \mathbb{R}^{D},
\end{align}
where $\mathbf{T}$ is modality-shared since temporal information is modality-agnostic.

\textbf{Masked Input.}
We set up learnable modality-specific mask embeddings $\mathbf{m}_{t_q}^{\mathrm{v}}$, $\mathbf{m}_{t_k}^{\mathrm{a}}$ to separately represent the masked image and action inputs.
Similarly, positional and timestep embeddings are added to the masked features:
\begin{align}
    \mathbf{m}_{t_q}^{\mathrm{v}^{'}} &= \mathbf{m}_{t_q}^{\mathrm{v}} + \mathbf{T}_{q} + \frac{1}{N}\sum_{n=1}^{N}\mathbf{E}_{n}^\mathrm{pos}, \ \ &\mathbf{m}_{t_q}^{\mathrm{v}} \in \mathbb{R}^{D},  \\
    \mathbf{m}_{t_k}^{\mathrm{a}^{'}} &= \mathbf{m}_{t_k}^{\mathrm{a}} + \mathbf{T}_{k}, \ \ &\mathbf{m}_{t_k}^{\mathrm{a}} \in \mathbb{R}^{D}, 
\end{align}
Finally, the features of the visible images and actions, along with the masked embeddings, are fed into the sequence-aware transformer, as shown in \cref{fig4:framework}.

\textbf{Target.}
For masked-out images, we use the exponential moving average (EMA) updated vision encoder $V_{\theta^{'}}^{\mathrm{EMA}}$ to obtain its patch features $\mathbf{H}_{t_q}^{\mathrm{v, gt}}\in\mathbb{R}^{N \times D}$. 
Then, we average these patch features to obtain the ground truth:
\begin{align}
    \mathbf{f}_{t_q}^\mathrm{v,gt} &=  \mathbf{T}_{q} + \frac{1}{N}\sum_{n=1}^{N}\mathbf{H}_{t_q, n}^\mathrm{v, gt}, \ \ \mathbf{f}_{t_q}^\mathrm{v,gt}\in\mathbb{R}^{D},
\end{align}
For actions that are masked out, we directly utilize the original action $\mathbf{a}_{t_k \rightarrow t_{k+1}}$ for supervision, as shown in \cref{fig4:framework}.

\textbf{Loss.}
The loss is calculated using the Smooth L1 Loss between the features predicted by our model and target features as follows:
\begin{align}
    \mathcal{L}_{\mathrm{total}}&= \mathcal{L}_{\mathrm{SmoothL1}}(\mathbf{f}_{t_q}^\mathrm{v,gt}, \mathbf{f}_{t_q}^{\mathrm{v'}}) + \mathcal{L}_{\mathrm{SmoothL1}}(\mathbf{a}_{t_k \rightarrow t_{k+1}}, \mathbf{a}_{t_k \rightarrow t_{k+1}}^{'}),
\label{eq:loss}
\end{align}
where $\mathbf{f}_{t_q}^{\mathrm{v'}}$ and $\mathbf{a}_{t_k \rightarrow t_{k+1}}^{'}$ are the masked visual feature and action predicted by the sequence-aware transformer $S_\psi$.

\subsection{Sampling Strategy}
\label{method:sampling}

\begin{figure*}[t!]
\centering
\includegraphics[width=2.1\columnwidth]{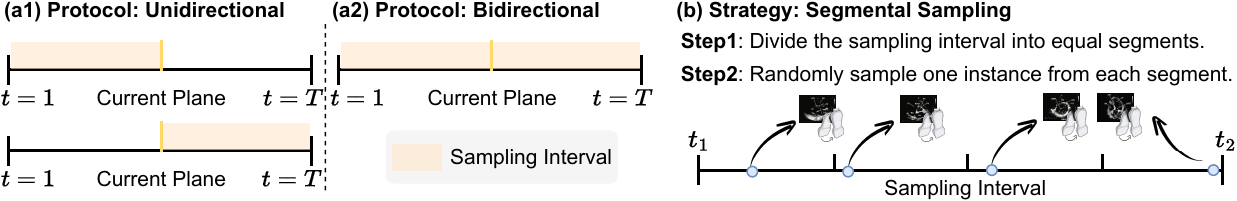}
\caption{
Diagram illustrating sampling protocol and strategy.
(a1) 
Unidirectional protocol samples from only one side of the current plane.
(a2)
Bidirectional protocol allows samples the entire scan, leveraging as much individual cardiac structural information as possible.
(b) 
Segmental sampling ensures that sampled instances span the entire interval.
}
\label{fig5:sampling}
\end{figure*}

In the following, we provide an explanation of our sampling strategy.

\textbf{Unidirectional protocol.}
Given an image $ I_{t_L} $ from a specific scan, a natural idea is to follow the scanning order of the sonographer, that is, to sample instances where $t < t_L$ in the same scan to construct a historical scanning trajectory.
In particular, due to variability in cardiac structures between individuals, sampling must be restricted to the same scan. 
Otherwise, mixing information from different individuals would prevent the model from learning individual-specific structural information.
Furthermore, from the perspective of symmetry, we also sampled from the other side to form a sequence during pre-training, as shown in \cref{fig5:sampling}(a1).

\textbf{Bidirectional protocol.}
In clinical practice, novice physicians typically begin by scanning relatively simple planes before transitioning to more challenging ones.
When handling complex planes, it is essential to gather as much information as possible about the patient's cardiac structure to assist in the scanning process.
Thus, we also consider a bidirectional sampling method, which captures the maximum amount of plane information (\cref{fig5:sampling} a2).

\textbf{Segmental sampling.}
The previous two methods define the sampling intervals, within which we apply the segmental sampling strategy~\cite{wang2016temporal}, as shown in \cref{fig5:sampling}(b).  
This method first divides the sampling interval into \( L-1 \) equal segments and then randomly samples one point from each segment with equal probability.  
This method ensures sampling coverage across the entire interval, enhancing the diversity of sampled planes to capture as much cardiac structural information as possible.

\begin{figure}[!t]
\centering
\includegraphics[width=1\columnwidth]{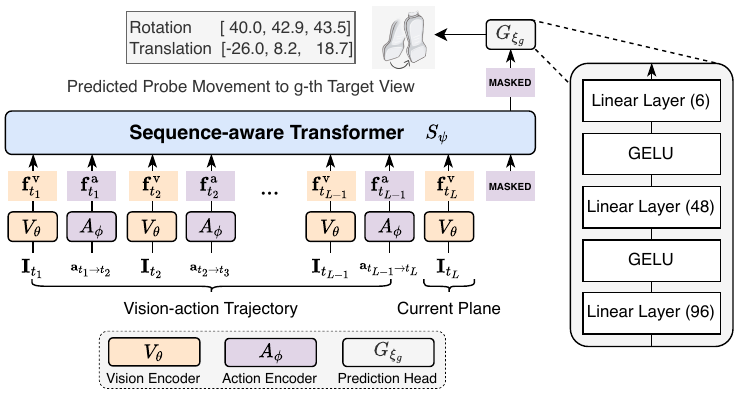}
\caption{
Fine-tuning for probe movement guidance.
The pre-trained sequence-aware transformer is fine-tuned with randomly initialized prediction heads.
}
\label{fig6:downstream}
\end{figure}

\subsection{Downstream Transfer}
\label{method:transfer}
After pre-training, the sequence-aware transformer $S_\psi$ is further fine-tuned for the probe movement guidance task.  
The goal of the probe movement guidance task is to predict the relative movement action from any arbitrary plane to the ten standard planes.  
During fine-tuning, we introduce ten guidance prediction heads $G_{\xi_{g}}, g \in \mathbb{Z}, \ 1 \leq g \leq 10$ to predict the action toward each standard plane, as illustrated in \cref{fig6:downstream}.  
All ten heads are trained simultaneously during fine-tuning, while during inference, the corresponding head is selected based on the target standard plane.
Meanwhile, the sequence-aware transformer is shared among the ten heads, as the cardiac structural information it learns is applicable for navigating toward all standard planes.
Similar to pre-training, given an image $ I_{t_L} $ from a specific scan, we sample $ L-1 $ instances to form a historical sequence.
Formally, the forward process for predicting action from $ I_{t_L} $ to $g$-th standard plane can be represented as:
\begin{align}
    m_{t_L}^{\mathrm{a'}} = S_\psi([&\mathbf{I}_{t_1}, \mathbf{a}_{t_1 \rightarrow t_2}, \cdots,  \mathbf{I}_{t_{L-1}}, \mathbf{a}_{t_{L-1} \rightarrow t_L}, \mathbf{I}_{t_L}, m_{t_L}^\mathrm{a}]),  \\
    & a_{t_L \rightarrow t_{g\text{-th standard plane}}} = G_{\xi_{g}}(m_{t_L}^\mathrm{a'}).
\end{align}

During fine-tuning, the historical sequence sampling protocol remains consistent with pre-training. 
In unidirectional protocol, when predicting the movement from $ I_{t_L} $ to standard planes with timestamp $ t > t_L $, the historical sequence is sampled from $ t \leq t_L $. 
Similarly, for standard planes with timestamp $ t < t_L $, the sequence is segmentally sampled from $ t \geq t_L $.
In bidirectional protocol, during training, the sampling process is the same as pre-training.
During evaluation, when predicting the movement from $ I_{t_L} $ to a specific standard plane, both the labeled standard plane and samples with similar visual features which are identified using a standard plane classifier are removed from the scan.
By filtering out these samples, information leakage during evaluation can be avoided.
Otherwise, if the target plane is present in the sequence, its known pose would allow direct derivation of the action, undermining the evaluation of the model’s true capability.
The historical sequence is then segmentally sampled from the remaining instances inside the scan.
Importantly, during evaluation, for both unidirectional and bidirectional protocols, the step-2 of segmental sampling (\cref{fig5:sampling}) is replaced with a fixed strategy that selects the sample at the center of each segment, thereby avoiding the introduction of randomness into the validation process.

\section{Experiments}
\label{sec:exp}
In this section, we provide thorough validation to demonstrate the method's effectiveness and include implementation details for reproducibility.

\subsection{Implementation Details}
\textbf{Dataset.}
The echocardiographic scanning dataset contains 476 demonstrations from routine clinical scans, with a total of 1.67 million image-pose pairs.
The dataset was collected using a GE Vivid E7 machine (\textit{General Electric, USA}) equipped with a M5S probe.
In the experiments, we divided the dataset into 284 scans (1.03 million sample pairs) for the training set, 72 scans (0.28 million sample pairs) for validation set 1, and 120 scans for validation set 2 (0.36 million sample pairs).
Validation set 1 includes 38 subjects with an age distribution of 41.1 ± 15.3 years (max: 69, min: 18). 
Among them, 9 subjects have cardiac diseases, including atrial septal defect, ventricular septal defect, degenerative valve changes, and ventricular septal hypertrophy. 
The oldest patient in this subgroup is 69 years old, with a mean age of 47.5 years.
Validation set 2 consists of 60 subjects with an age distribution of 54.2 ± 10.7 years (max: 75, min: 26). 
The cohort was divided into three age groups: under 50 years, 50 to 60 years, and over 60 years, with each group containing 20 subjects and maintaining a 1:1 male-to-female ratio. 
Unless otherwise specified, experiments were conducted on validation set 1.
Please note that the validation set comprises individuals not encountered during training, allowing for an assessment of the algorithm's generalization capabilities.
The whole data collecting process was approved and supervised by the
Tsinghua University Medical Ethics Committee (Approval No.THU-01-2025-1001). 
A signed informed consent form was collected from each participant.

\textbf{Model architecture.}
We use a ViT-Small/16 model as the default vision encoder $V_\theta$ and target encoder $V_{\theta^{'}}^{\mathrm{EMA}}$.
The action encoder $A_\phi$ is a single-layer MLP that maps the original 6-DOF actions to a 192-dimensional action space.
To match the dimensionality of visual and action features in the sequence-aware transformer, we add a MLP layer to project the visual features to 192 dimensions.
The sequence-aware transformer is a custom architecture with 4 transformer layers (hidden size 192, 4 attention heads, and MLP ratio 2).
All models are implemented in Pytorch framework.

\textbf{Pre-training.} 
The vision encoder is firstly pre-trained on training set using the I-JEPA method~\cite{assran2023self} with their default hyper-parameters.
During our pre-training, the vision encoder is further jointly trained with the sequence-aware transformer.
The training batch size is 512. 
The model is optimized using AdamW for 50 epochs, with the first 7 epochs designated for warm-up, followed by a cosine learning rate schedule starting at an initial learning rate of 0.00025.
Data augmentation only involves random resized crop, with a crop size of 224$\times$224.
The mask ratio is set to range between 0.3 and 0.5.
The default sequence length $L$ is set to 6 unless otherwise specified.

\textbf{Downstream fine-tuning.}
For fine-tuning, after loading the pre-trained parameters, the model is optimized using AdamW with a batch size of 512 and a learning rate of $ 1 \times 10^{-4} $. 
A cosine learning rate schedule is employed, and the training is conducted for 5 epochs.
For evaluation protocols (1)/(2), the vision encoder is jointly fine-tuned with other modules.
Additionally, for evaluation protocol (2), we apply a drop path rate of 0.2 to the sequence-aware transformer and a layer-wise learning rate decay of 0.4 to the vision encoder.
The baselines use the same settings.
For evaluation protocol (3), we empirically find that freezing the vision encoder achieves better performance.
All experiments are run on the Ubuntu system equipped with 8 NVIDIA RTX 3090 GPUs.

\textbf{Downstream evaluation.}
We evaluate our pre-trained model under three protocols as follows.
(1) Single-frame protocol: The input consists only of the current plane's ultrasound image, without any past sequences. 
The network architecture includes the vision encoder, which loads our pre-trained weights, and prediction heads.
This setup allows us to verify that the vision encoder learns better features during sequence-aware pre-training.
(2) Sequential and unidirectional protocol: The input consists of the current plane's ultrasound image and the historical scanning sequence, sampled following the method mentioned in \cref{method:transfer}.
Prediction heads are added to the pre-trained network architecture, and both the vision encoder and sequence-aware transformer are initialized with the pre-trained parameters.
(3) Sequential and bidirectional protocol: Compared to unidirectional protocol, the only difference lies in the sampling method, as described in \cref{method:transfer}.
The default protocol is sequential and bidirectional unless otherwise specified.

\textbf{Evaluation metric.}
The metric used is the Mean Absolute Error (MAE) between the predicted probe movement action $\mathbf{a}^{'}$ and the ground truth $\mathbf{a}$.
\begin{align}
    \text{Translation MAE} &= \frac{1}{3} \sum_{i=1}^{3} |\mathbf{a}_i - \mathbf{a}^{'}_i|, \\
    \text{Rotation MAE} &= \frac{1}{3} \sum_{i=4}^{6} |\mathbf{a}_i - \mathbf{a}^{'}_i|.
\end{align}
Unless specified otherwise, we report the average MAE across the ten planes.

\textbf{Baselines.}
We compare a series of pre-training methods, including the echocardiogram foundation model EchoCLIP~\cite{christensen2024vision}, the universal ultrasound foundation model USFM~\cite{jiao2024usfm}, and advanced self-supervised learning algorithms pre-trained on our dataset, \textit{i.e.,} US-MAE~\cite{he2022masked}, US-IJEPA~\cite{assran2023self}, and US-MoCo~\cite{chen2021empirical}.  
Additionally, we evaluate medical vision models such as BioMedCLIP~\cite{zhang2023biomedclip} and LVM-Med~\cite{mh2024lvm}, as well as general vision models like DeiT~\cite{touvron2021training} and DINOv2~\cite{oquab2023dinov2}.  
Furthermore, we compare a sequence probe movement guidance model for obstetric ultrasound, US-GuideNet~\cite{droste2020automatic}, and a classic sequence decision model, Decision-Transformer~\cite{chen2021decision}.
It is worth noting that US-GuideNet, like our model, uses a vision transformer as the vision encoder rather than the originally proposed mobilenetv2~\cite{sandler2018mobilenetv2}.
Additionally, both US-GuideNet and the Decision Transformer utilize vision encoders initialized with pre-trained parameters from US-IJEPA, making them two strong baselines.
Under the sequential protocol, the baseline named ``Scratch" refers to the results obtained by training our architecture from scratch.
In the subsequent ablation experiments, the vision encoder in the w/o sequence-aware pre-training baseline is also initialized with pre-trained parameters from US-IJEPA, rather than being trained from scratch.

\begin{table*}[!t]\normalsize
\caption{
\textbf{Comparison with baseline methods.} 
We report mean absolute error (lower is better) on the ten most common standard planes.
“Trans.” and “Rot.” represent the translation (mm) and rotation (°) errors.
\dag~indicates that our pre-trained vision encoder and initialized prediction heads are used.
\ddag~indicates that these methods use the same vision encoder as our method, initialized with US-IJEPA pre-trained weights.
}
\vspace{-10pt}
\label{tab1:baselines}
\begin{center}
\resizebox{2.1\columnwidth}{!}{
\begin{tabular}{p{2.8cm}*{22}{P{0.6cm}}}
\toprule
    \multirow{2}{*}{Method} & \multicolumn{2}{c}{\footnotesize PLAX} & \multicolumn{2}{c}{\footnotesize PSAX-AV} & \multicolumn{2}{c}{\footnotesize PSAX-PV} & \multicolumn{2}{c}{\footnotesize PSAX-MV} & \multicolumn{2}{c}{\footnotesize PSAX-PAP} & \multicolumn{2}{c}{\footnotesize PSAX-APEX}  & \multicolumn{2}{c}{\footnotesize A4C} & \multicolumn{2}{c}{\footnotesize A5C} & \multicolumn{2}{c}{\footnotesize A3C} & \multicolumn{2}{c}{\footnotesize A2C} &\multicolumn{2}{c}{Average}\\
    & {\footnotesize Trans.} & {\footnotesize Rot.} & {\footnotesize Trans.} & {\footnotesize Rot.}& {\footnotesize Trans.} & {\footnotesize Rot.}& {\footnotesize Trans.} & {\footnotesize Rot.}& {\footnotesize Trans.} & {\footnotesize Rot.}& {\footnotesize Trans.} & {\footnotesize Rot.}& {\footnotesize Trans.} & {\footnotesize Rot.}& {\footnotesize Trans.} & {\footnotesize Rot.}& {\footnotesize Trans.} & {\footnotesize Rot.}& {\footnotesize Trans.} & {\footnotesize Rot.}  & {\footnotesize Trans.} & {\footnotesize Rot.} \\
\midrule
\midrule
    \multicolumn{21}{l}{\textit{Single-frame protocol, not pre-trained on ultrasound data}} &  \\
    Scratch \cite{dosovitskiy2020image} & 8.84 &  7.78 & 8.12 & 8.31 & 8.65 & 9.02 & 8.18 & 9.21 & 8.05 & 9.13 & 9.56 & 9.89 & 8.86 & 7.95 & 9.06 & 10.04 & 9.35 & 10.22 & 9.09 & 11.97 & 8.78 & 9.35 \\
    DeiT \cite{touvron2021training}& 8.50 & 7.27 & 8.09 & 8.02 & 8.51 & 8.71 & 7.90 & 8.66 & 7.85 & 8.62 & 9.21 & 9.23 & 8.43 & 7.37 & 8.55 & 9.54 & 8.70 & 9.71 & 8.52 & 11.28 & 8.43 & 8.84 \\
    DINOv2 \cite{oquab2023dinov2}& 8.31 & 7.21 & 7.82 & 7.82 & 8.36 & 8.75 & 7.70 & 8.66 & 7.66 & 8.42 & 9.09 & 9.04 & 8.40 & 7.27 & 8.52 & 9.41 & 8.70 & 9.38 & 8.62 & 11.27 & 8.32 & 8.72 \\
    \midrule
    \multicolumn{21}{l}{\textit{Single-frame protocol, pre-trained on ultrasound data}} & \\
    BioMedCLIP \cite{zhang2023biomedclip}& 8.40 & 7.33 & 7.89 & 8.00 & 8.44 & 8.84 & 7.79 & 8.88 & 7.78 & 8.85 & 9.16 & 9.38 & 8.62 & 7.81 & 8.73 & 10.04 & 8.88 & 9.59 & 8.75 & 11.62 & 8.44 & 9.03 \\
    LVM-Med \cite{mh2024lvm}& 8.55 & 7.26 & 8.00 & 7.94 & 8.42 & 8.62 & 7.95 & 8.81 & 7.89 & 8.85 & 9.31 & 9.27 & 8.70 & 7.59 & 8.88 & 9.65 & 9.07 & 9.65 & 8.80 & 11.34 & 8.56 & 8.90 \\
    US-MoCo \cite{chen2021empirical} & 8.75 & 7.46 & 8.08 & 7.90 & 8.29 & 8.64 & 7.95 & 8.76 & 7.90 & 8.64 & 9.22 & 9.23 & 8.55 & 7.40 & 8.78 & 9.76 & 8.88 & 9.70 & 8.74 & 11.55 & 8.51 & 8.90 \\
    US-IJEPA \cite{assran2023self} & 8.28 & 7.24 & 7.82 & 7.75 & 8.36 & 7.98 & 7.74 & 8.73 & 7.74 & 8.46 & \underline{8.84} & 9.06 & 8.46 & 7.57 & 8.55 & 9.72 & 9.01 & 9.42 & 8.67 & \underline{10.81} & 8.35 & 8.67 \\
    US-MAE \cite{he2022masked} & 8.31 & 7.11 & \underline{7.74} & 7.61 & 8.30 & \underline{8.23} & 7.69 & 8.55 & 7.70 & 8.56 & 9.02 & 9.05 & 8.35 & 7.26 & 8.47 & 9.67 & 8.55 & 9.46 & 8.45 & 11.05 & 8.26 & 8.66 \\
    USFM \cite{jiao2024usfm}& 8.34 & 7.14 & \underline{7.74} & 7.69 & \underline{8.28} & 8.51 & 7.62 & \underline{8.38} & 7.69 & \underline{8.41} & 9.02 & \underline{9.00} & 8.20 & 7.18 & 8.33 & \underline{9.38} & \underline{8.51} & 9.37 & 8.39 & 11.14 & 8.21 & 8.62 \\
    EchoCLIP \cite{christensen2024vision}& \underline{8.29} & \textbf{6.86} & 7.78 & \underline{7.52} & 8.45 & 8.51 & \underline{7.53} & 8.44 & \underline{7.60} & 8.47 & 9.17 & 9.18 & \underline{8.16} & \underline{7.03} & \underline{8.22} & \textbf{9.31} & 8.58 & \textbf{8.88} & \underline{8.35} & 10.97 & 8.21 & 8.52 \\
    \rowcolor{gray!20} $\textbf{Ours}^{\dag}$ & \textbf{8.00} & \underline{6.90} & \textbf{7.59} & \textbf{7.25} & \textbf{8.12} & \textbf{7.65} & \textbf{7.38} & \textbf{8.08} & \textbf{7.44} & \textbf{8.00} & \textbf{8.63} & \textbf{8.58} & \textbf{7.94} & \textbf{6.99} & \textbf{8.01} & 9.50 & \textbf{8.33} & \underline{9.11} & \textbf{8.08} & \textbf{10.66} & \textcolor{BrickRed}{\textbf{7.95}} & \textcolor{BrickRed}{\textbf{8.27}} \\
\midrule
    \multicolumn{21}{l}{\textit{Sequential, unidirectional protocol, pre-trained on ultrasound data}} & \\
    Scratch & 7.83 & 6.52 & 7.56 & 7.54 & 8.12 & 7.60 & 7.05 & 7.94 & 7.22 & 7.91 & \underline{8.06} & 8.49 & 7.61 & 6.71 & 7.69 & 8.86 & 7.79 & 8.72 & 7.70 & 11.21 & 7.66 & 8.15 \\
    $\text{US-GuideNet}^{\ddag}$ \cite{droste2020automatic} 
    & \textbf{7.32} & 6.14 & \underline{6.97} & \underline{6.98} & \underline{7.51} & 7.07 & 6.93 & \underline{7.66} & 6.97 & 7.61 & 8.16 & 8.31 & 7.35 & 6.35 & 7.48 & 8.95 & 7.88 & 8.50 & 7.60 & 10.53 & 7.42 & 7.81 \\
    $\text{Decision-T}^{\ddag}$ \cite{chen2021decision} & 7.51 & \underline{6.12} & 6.97 & 7.17 & 7.53 & \underline{7.06} & \underline{6.78} & \underline{7.66} & \underline{6.96} & \underline{7.56} & 8.21 & \underline{8.17} & \underline{7.03} & \underline{6.23} & \underline{7.18} & \underline{8.81} & \underline{7.68} & \textbf{8.02} & \underline{7.41} & \underline{10.49} & 7.33 & 7.73\\
    \rowcolor{gray!20} \textbf{Ours} & \underline{7.44} & \textbf{6.06} & \textbf{6.94} & \textbf{6.86} & \textbf{7.44} & \textbf{6.90} & \textbf{6.34} & \textbf{7.30} & \textbf{6.55} & \textbf{7.35} & \textbf{7.46} & \textbf{8.05} & \textbf{6.79} & \textbf{6.21} & \textbf{6.83} & \textbf{8.33} & \textbf{7.26} & \underline{8.04} & \textbf{7.04} & \textbf{10.40} & \textcolor{BrickRed}{\textbf{7.01}} & \textcolor{BrickRed}{\textbf{7.55}} \\
\midrule
    \multicolumn{21}{l}{\textit{Sequential, bidirectional protocol, pre-trained on ultrasound data}} & \\
    Scratch & 6.50 & 5.75 & 5.94 & 6.13 & 6.38 & 6.39 & 5.49 & 6.94 & 5.39 & 7.10 & \underline{6.96} & 6.98 & 6.18 & 5.73 & 7.21 & 8.31 & 7.38 & 8.45 & 6.85 & 9.61 & 6.43 & 7.14 \\
    $\text{US-GuideNet}^{\ddag}$ \cite{droste2020automatic}& 6.26 & 5.71 & 5.48 & 6.10 & 6.02 & \underline{6.19} & 5.39 & \underline{6.39} & \underline{5.34} & 6.21 & 7.02 & 7.22 & \underline{5.85} & \underline{5.58} & \underline{6.00} & \underline{8.18} & \underline{6.24} & \underline{7.46} & 6.11 & \underline{9.43} & 5.97 & 6.85\\
    $\text{Decision-T}^{\ddag}$ \cite{chen2021decision} & \textbf{6.12} & \underline{5.54} & \underline{5.39} & \textbf{5.71} & \underline{5.87} & 6.23 & \underline{5.18} & 6.40 & 5.36 & \underline{6.11} & 7.08 & \textbf{6.86} & 6.00 & 5.62 & 6.01 & 8.22 & 6.39 & 7.47 & \underline{6.09} & 9.89 & 5.95 & 6.81 \\
    \rowcolor{gray!20} \textbf{Ours} & \underline{6.16} & \textbf{5.31} & \textbf{5.01} & \underline{5.86} & \textbf{5.86} & \textbf{5.98} & \textbf{4.76} & \textbf{6.02} & \textbf{4.84} & \textbf{5.82} & \textbf{6.33} & \underline{6.93} & \textbf{4.43} & \textbf{4.91} & \textbf{4.59} & \textbf{7.40} & \textbf{4.95} & \textbf{6.80} & \textbf{4.32} & \textbf{8.35} & \textcolor{BrickRed}{\textbf{5.13}} & \textcolor{BrickRed}{\textbf{6.34}}\\ 
\bottomrule
\end{tabular}
}
\end{center}
\end{table*}

\begin{table*}[t!]
\caption{Statistical significance testing of our model performance vs. the best baselines.}
\vspace{-10pt}
\begin{center}
\resizebox{1.5\columnwidth}{!}{
\begin{tabular}{l|cc|cc|cc}
    \toprule
\multirow{2}{*}{Method} & \multicolumn{2}{c|}{Bidirectional} & \multicolumn{2}{c|}{Unidirectional} & \multicolumn{2}{c}{Single-frame}\\
& Trans. & Rot. & Trans. & Rot. & Trans. & Rot.\\
    \midrule
Paired t-test & p$<$0.0001 & p$<$0.0001 & p$<$0.0001 & p$<$0.0001 & p$<$0.0001 & p$<$0.0001 \\
Bootstrap (95\% CI) & [0.64, 0.65] & [0.63, 0.64] & [0.23, 0.24] & [0.19, 0.20] & [0.38, 0.43] & [0.24, 0.29]\\
    \bottomrule
\end{tabular}
}
\end{center}
\label{tab:stat_test}
\end{table*}

\subsection{Comparison with Baselines}
Firstly, under the single-frame protocol, our method achieves lower prediction error compared to the echocardiogram interpretation foundation model EchoCLIP~\cite{christensen2024vision}, which was trained with over 1 million echocardiogram video-text pairs.
Our method also surpasses the universal ultrasound foundation model USFM~\cite{jiao2024usfm}, which was trained on a large-scale dataset containing over 2 million ultrasound images from 12 common organs.
Secondly, under the sequential protocol, a significant reduction in probe movement prediction error is observed, indicating that the historical scanning sequence contains valuable information about the individual’s cardiac structure, which benefits subsequent scans.
Additionally, compared to the obstetric ultrasound guidance algorithm US-GuideNet~\cite{droste2020automatic} and the classic sequential decision-making method Decision-Transformer~\cite{chen2021decision}, our method achieves superior performance.
Better performance indicates that our method, after pretraining, is more effective at extracting individual-specific cardiac structural information from the sequence.
The bidirectional protocol outperforms the unidirectional one as it samples from a larger interval, capturing more diverse spatial pose information from different planes, thereby enabling better modeling of individual cardiac structures.

To more rigorously validate the performance improvement of our model, we conducted statistical significance tests comparing our model’s performance against that of the best baseline across three protocols.
In \cref{tab:stat_test}, the paired t-test results show that the mean performance of our model is significantly lower than that of the baselines, while the bootstrap resampling further confirms a statistically significant improvement. 
Based on the above results, our model indeed shows a clear improvement over the baselines.

\begin{table*}[t!]\large
\caption{Performance analysis based on demographic under bidirectional protocol.}
\vspace{-10pt}
\begin{center}
\resizebox{2.1\columnwidth}{!}{
\begin{tabular}{l|cc|cc|cc|cc|cc|cc}
    \toprule
\multirow{2}{*}{Characteristics} & \multicolumn{2}{c|}{Cardiac disease} & \multicolumn{2}{c|}{Sex-male} & \multicolumn{2}{c|}{Sex-female} & \multicolumn{2}{c|}{Age$<$50} & \multicolumn{2}{c|}{50$\leq$Age$\leq$60} &  \multicolumn{2}{c}{Age$>$60}\\
& Trans. & Rot. & Trans. & Rot. & Trans. & Rot. & Trans. & Rot. & Trans. & Rot. & Trans. & Rot.\\
    \midrule
Decision-T & 5.56 & 6.92 & 7.46 & 8.20 & 5.87 & 8.00 & 6.63 & 8.53 & 5.76 & 7.86 & 7.66 & 7.98 \\ 
\rowcolor{gray!20} \textbf{Ours} & \textbf{5.23} & \textbf{6.13} & \textbf{6.73} & \textbf{7.81} & \textbf{5.47} & \textbf{7.56} & \textbf{6.20} & \textbf{7.81} & \textbf{5.46} & \textbf{7.83} & \textbf{6.71} & \textbf{7.48} \\
    \midrule
Paired t-test & p$<$0.0001 & p$<$0.0001 & p$<$0.0001 & p$<$0.0001 & p$<$0.0001 & p$<$0.0001 & p$<$0.0001 & p$<$0.0001 & p$<$0.0001 & p$<$0.0001 & p$<$0.0001 & p$<$0.0001 \\
Bootstrap (95\% CI) & [0.32, 0.33] & [0.78, 0.80] & [0.72, 0.73] & [0.38, 0.40] & [0.39, 0.41] & [0.44, 0.45] & [0.42, 0.43] & [0.71, 0.73] & [0.29, 0.30] & [0.02, 0.04] & [0.94, 0.95] & [0.49, 0.51] \\
    \bottomrule
\end{tabular}
}
\end{center}
\label{tab:demographi_stat_test}
\end{table*}

\subsection{Analysis on Demographic Variables}
Cardiac structure is influenced by multiple demographic variables, such as disease status, sex, and age. 
These factors may further affect ultrasound imaging appearances and thereby introduce additional challenges to model perception and decision-making.
Therefore, we conducted validation analyses on these potential influencing factors.

First, regarding disease conditions, validation set 1 includes nine subjects with cardiac diseases including atrial septal defect, ventricular septal defect, degenerative valve changes, and ventricular septal hypertrophy. 
The oldest subject in this group is 69 years old, with a mean age of 47.5.
As shown in \cref{tab1:baselines} and \cref{tab:stat_test}, our model outperformed the baseline model on validation set 1 containing patients. 
Furthermore, we tested the nine patient cases as an independent validation set.
The results presented in the "cardiac disease" column of \cref{tab:demographi_stat_test} demonstrate that our model maintains stable performance while still surpassing the baseline model.

Second, we compared the performance of our proposed model with the best baseline, Decision-T, across sex and age groups on validation set 2. 
This validation set was specifically constructed to balance sex and age distributions, consisting of 60 subjects with an age distribution of 54.2 ± 10.7 years (max: 75, min: 26). The subjects were divided into three age groups: under 50 years, 50 to 60 years, and over than 60 years, with 20 subjects per group and a 1:1 male-to-female ratio in each. 
This grouping strategy ensures gender balance and a broader age coverage, thereby better reflecting the demographic characteristics of real clinical populations.
In terms of sex, our model outperformed the baseline in all cases. 
Both models performed better in the female group, which may be attributed to the fact that approximately one-third of the male subjects were smokers—smoking can lead to lung deposits that impair imaging clarity and pose additional challenges for model perception. 
Overall, our model demonstrated effectiveness across both sexes and surpassed the current best model.
Across age groups, our model outperformed or matched the baseline. 
Although some performance fluctuations were observed across age groups, these occurred in both our model and the baseline model with similar trends, likely due to inherent variations in data distribution within each subgroup. 
Nevertheless, our model consistently demonstrated effectiveness across all age groups and overall outperformed the current best model.

\begin{table*}[t!]
\caption{Ablation study on sequence-aware pre-training.}
\vspace{-10pt}
\begin{center}
\resizebox{1.5\columnwidth}{!}{
\begin{tabular}{c!{\vrule width 1.25pt}c|ll!{\vrule width 1.25pt}c|ll}
    \toprule
Pre-train & \multicolumn{1}{c|}{Protocol} &  Trans.(mm) $\downarrow$ & Rot.(°) $\downarrow$ & \multicolumn{1}{c|}{Protocol} & Trans.(mm) $\downarrow$ & Rot.(°) $\downarrow$ \\
    \midrule
\XSolidBrush &\multirow{2}{*}{Unidirectional} & 7.41 & 7.90  & \multirow{2}{*}{Bidirectional} & 5.53 & 6.69 \\
\Checkmark & & \textbf{7.01} \textbf{\textcolor{mygreen}{(-5.4\%)}} & \textbf{7.55} \textbf{\textcolor{mygreen}{(-4.4\%)}} & & \textbf{5.12} \textbf{\textcolor{mygreen}{(-7.4\%)}} & \textbf{6.34} \textbf{\textcolor{mygreen}{(-5.2\%)}} \\
    \bottomrule
\end{tabular}
}
\end{center}
\label{tab:ab_seqaware_pretrain}
\end{table*}

\subsection{Ablation and Discussion of Key Components}
In this section, we perform ablation studies to evaluate the effectiveness of each proposed component.

\begin{figure*}[t!]
\centering
\includegraphics[width=1.5\columnwidth]{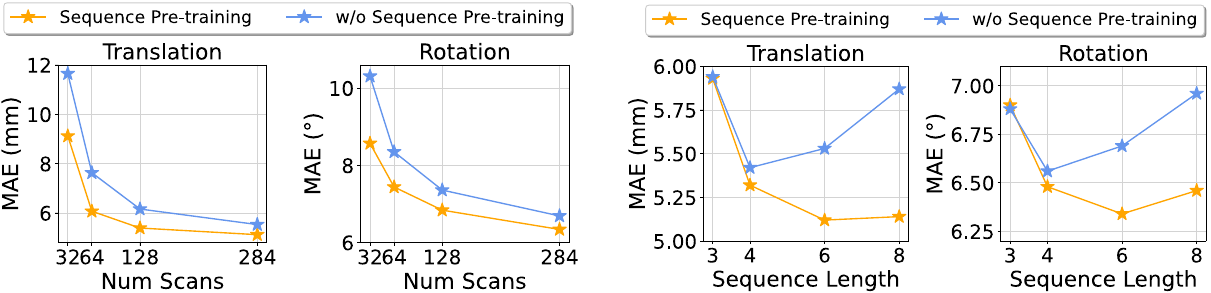}
\caption{
Ablation study under various training scanning demonstrations and sequence length.
}
\label{fig:num_scan_and_seq_length}
\end{figure*}

\textbf{Sequence-aware pre-training.}
Although prior work~\cite{droste2020automatic} also employed sequence modeling, it alone cannot automatically extract cardiac structural information, potentially wasting valuable sequence data.
As shown in \cref{fig:num_scan_and_seq_length} (right), conventional sequence models fail to deliver performance gains when sequence length exceeds 4.
In contrast, sequence-aware pre-training, through the masked prediction paradigm, requires the model to fully grasp the causal relationships between planes' visual feature and probe movement actions.
After pre-training, the model acquires the following capabilities: (1) Given a plane's ultrasound image and a probe movement action, it can predict the image features of the resulting plane after the adjustment; (2) Given two planes' images, it can predict the relative movement action (relative spatial pose) between them.
In other words, the model develops a strong understanding of the heart's three-dimensional structure.
As shown in \cref{tab:ab_seqaware_pretrain}, the probe movement guidance error is reduced after sequence-aware pre-training.
Furthermore, in \cref{fig:num_scan_and_seq_length} (left), we demonstrate that sequence-aware pretraining improves guidance performance across all training scales.

\begin{figure*}[t!]
\centering
\includegraphics[width=1.5\columnwidth]{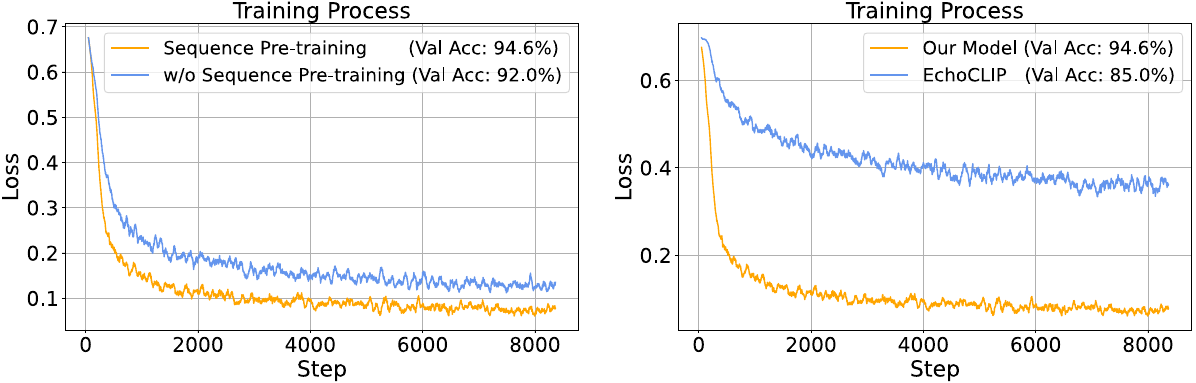}
\caption{
Ablation of pre-training (left) and sequential modeling (right) in learning personalized features.
}
\label{fig:personalized_cardiac_feature_ablation}
\end{figure*}

To further validate that sequence-aware pre-training helps in learning personalized features, we design a binary classification task to distinguish whether two inputs come from the same individual or different individuals.
During training, except for the classification head, both models are initialized with their respective optimal fine-tuned weights and frozen.
This allows us to verify how much individual-specific information is contained in the features learned by the model.
As shown in \cref{fig:personalized_cardiac_feature_ablation} (left), the pre-trained model exhibits a steeper loss decline during training and achieves better validation accuracy, indicating its ability to extract more personalized features.

\begin{table}[!t]
\caption{Ablation study on sequence modeling. }
\vspace{-10pt}
\begin{center}
\resizebox{1\columnwidth}{!}{
\begin{tabular}{c|p{2.9cm} p{2.9cm}}
    \toprule
\multicolumn{1}{c|}{Protocol} & Trans.(mm) $\downarrow$ & Rot.(°) $\downarrow$ \\
    \midrule
\multicolumn{1}{c|}{Single-frame} & 7.95 & 8.27\\
    \midrule
Sequential Unidirectional & 7.01 & 7.55 \\
    \midrule
Sequential Bidirectional & \textbf{5.12} \textbf{\textcolor{mygreen}{(-35.6\%)}} & \textbf{6.34} \textbf{\textcolor{mygreen}{(-23.3\%)}}\\
    \bottomrule
\end{tabular}
}
\end{center}
\label{tab:ab_seq_modeling}
\end{table}

\textbf{Sequential modeling.}
The decision-making paradigm based solely on a single image struggles to address individual variability due to the lack of effective information about personalized cardiac structures.
In contrast, the sequence modeling approach effectively leverages acquired structural information, leading to more accurate action decisions.
To validate the above hypothesis, both the baseline and our method are tested without pre-training, using the same model architecture. The only difference is that the baseline predicts based solely on a single image, while our method makes predictions based on sequence input.
As shown in the \cref{tab:ab_seq_modeling}, introducing sequence modeling significantly reduces prediction errors. 
Translation errors decrease by 35.6\%, while rotation errors decrease by a minimum of 23.3\%, strongly demonstrating the effectiveness of the sequence modeling paradigm.

To further validate that sequence modeling helps in learning personalized features, we also conduct the binary classification task to distinguish whether two inputs come from the same individual or different individuals.
For comparison, we choose the best-performing single-frame baseline model EchoCLIP.
As shown in \cref{fig:personalized_cardiac_feature_ablation} (right), our model converges faster and achieves higher accuracy on the validation set, indicating that it has learned more personalized features.

\begin{figure*}[t!]
\centering
\includegraphics[width=1.5\columnwidth]{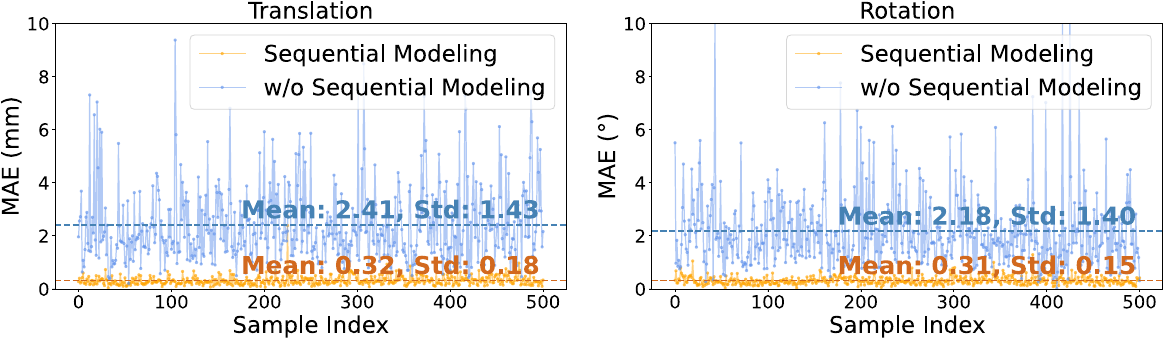}
\caption{
Robustness of sequential modeling to cardiac cycle variations.
}
\label{fig:ablation_cardiac_cycle}
\end{figure*}

Furthermore, since the heart undergoes continuous beating, which affects the content of cardiac ultrasound images, we further investigate the impact of the cardiac cycle on the model. 
Specifically, we sample 500 pairs of images from the same view but at different cardiac cycle phases, generate guidance actions toward ten standard views, and compute the average absolute error/difference between outputs from different cardiac phases. 
For the sequence modeling approach, we keep the same $L-1$ historical views unchanged and only varied the current input's cardiac phase. 
As shown in \cref{fig:ablation_cardiac_cycle}, the sequential modeling method significantly improves robustness against cardiac cycle variations, whereas the single-image-based approach is heavily influenced by the cardiac phase.

\textbf{Sequence length.}
We evaluate sequence-aware pretraining using different sequence lengths, specifically ranging from three to eight.
From \cref{fig:num_scan_and_seq_length} (right), we observe that our method outperforms the non-pretrained baseline when sequence length is greater than three.
Moreover, as the sequence length increases, the advantage of out pre-training expands, which aligns with our hypothesis that longer sequences contain more personalized information.
Meanwhile, we also observe that as the sequence length exceeds four, without sequence-aware pre-training, the guidance prediction error even exhibits a steep upward trend.
This indicates that capturing and modeling cardiac structure-related information from sequences is non-trivial and requires dedicated training.
This ablation also demonstrates that during pre-training, the model acquired the ability to capture personalized cardiac structure features in long sequences, which significantly improves the performance in the downstream task.

We also observe that in our method, as the sequence length increases, the rate of error reduction begins to decrease, and the error even increases at $ T = 8 $. 
This may suggests that the information in longer sequences starts to become redundant or even noisy.
In practical applications, sequences can be very long, and simply feeding all past sequences into the model may not yield performance gains while significantly increasing computational costs.
Therefore, a valuable direction would be to design more efficient sequence sampling algorithms, aiming to select samples that provide the maximum information gain as the sequence length increases.

\begin{table*}[!t]
\caption{Ablation study on sampling strategies. MAE values are averages of ten planes.}
\vspace{-10pt}
\begin{center}
\resizebox{1.5\columnwidth}{!}{
\begin{tabular}{c!{\vrule width 1.25pt}c|ll!{\vrule width 1.25pt}c|ll}
    \toprule
Strategy & Protocol & Trans.(mm) $\downarrow$ & Rot.(°) $\downarrow$ & Protocol & Trans.(mm) $\downarrow$ & Rot.(°) $\downarrow$ \\
    \midrule
Dense & \multirow{3}{*}{Unidirectional} & 7.59 & 8.07 & \multirow{3}{*}{Bidirectional} & 6.23 & 6.81 \\
Random & & 7.36 & 7.87 & & 5.70 & 6.81 \\
Segmental & & \textbf{7.01} \textbf{\textcolor{mygreen}{(-4.8\%)}} & \textbf{7.55} \textbf{\textcolor{mygreen}{(-4.1\%)}}& & \textbf{5.12} \textbf{\textcolor{mygreen}{(-10.2\%)}} & \textbf{6.34} \textbf{\textcolor{mygreen}{(-6.9\%)}}\\
    \bottomrule
\end{tabular}
}
\end{center}
\label{tab:ab_sample_strategy}
\end{table*}

\textbf{Sampling strategy.}
Since the scanning process generates long historical trajectories, selecting key instances from these trajectories is crucial for the model's performance.  
In \cref{tab:ab_sample_strategy}, we compare segmental sampling with two common sampling strategies: dense sampling~\cite{tran2015learning} and random sampling.
Dense sampling selects a sequence of $ L-1 $ consecutive frames preceding the current image, while random sampling selects $ L-1 $ frames randomly from the sampling interval to form a sequence.
From the table, it can be observed that under both protocols, segmental sampling achieves the best performance, while dense sampling performs the worst, and random sampling is slightly better than dense sampling.
Dense sampling performs poorly because the sampled interval is too localized, and the variation in probe poses within such a local interval is typically small. 
This means that the sequence contains highly localized cardiac structural information around the current plane, failing to provide global structural insights which is useful for navigating to other planes.
Meanwhile, the sequences obtained through random sampling can cover a larger sampling interval, resulting in a greater variety of probe poses (spanning multiple different planes), which also means more diverse structural information.
However, due to its randomness, random sampling may also face the same issue as dense sampling, where the samples could be too localized.
Unlike the previous two methods, segmental sampling first divides the sampling interval into $ L-1 $ equal segments and then randomly samples one instance from each segment. 
This approach prevents samples from being confined to localized intervals, which in turn leads to an improvement in overall performance.

\begin{figure*}[t!]
\centering
\includegraphics[width=1.5\columnwidth]{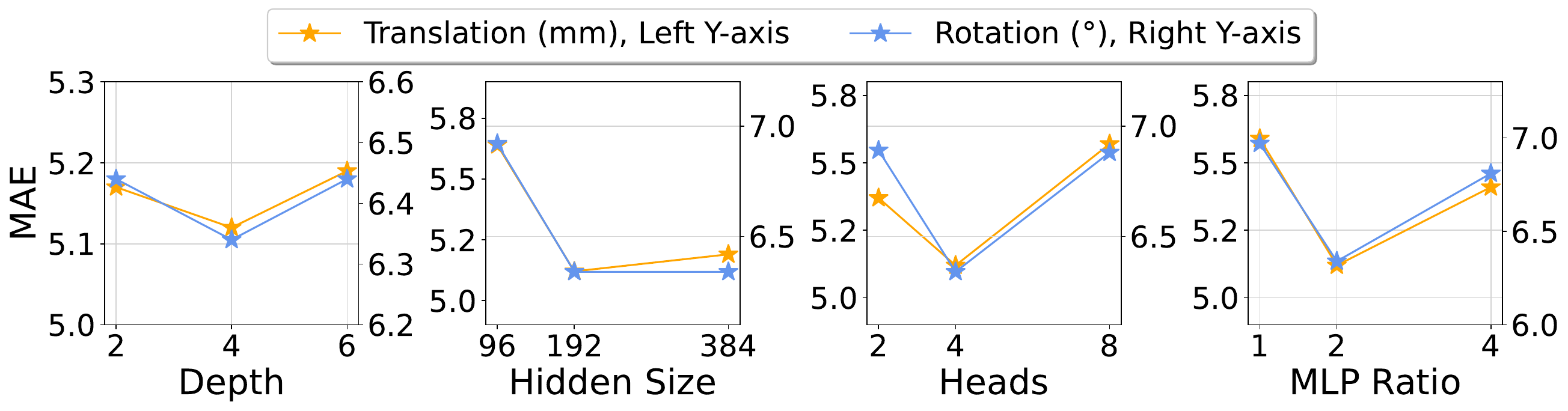}
\caption{
Ablations of sequence-aware transformer architecture.
}
\label{fig:ablation_of_mask_type_ratio}
\end{figure*}

\textbf{Sequence-aware transformer architecture.}
We further ablate the impact of structural hyperparameters, including depth, hidden size, number of heads, and MLP ratio, on the model's performance.
When ablating a specific dimension, the other dimensions use the default values, \textit{i.e.,} depth=4, hidden size=192, heads=4, and mlp ratio=2.
A trend of error first decreasing and then increasing is observed across all dimensions.
Except for the heads dimension, when the values of other dimensions decrease, the model's parameter also decreases, leading to a reduction in the model's expressive power.
When the values become too large, the model’s parameter scale may exceed the dataset size, which could lead the model to memorize the data rather than learn from it, thereby reducing its generalization ability.
As mentioned in \cite{vaswani2017attention}, multi-head attention allows the model to jointly attend to information from different representation subspaces.
Thus, reducing the number of heads decreases the model's expressive power.
Additionally, when the hidden size is fixed and the number of heads increases, the hidden size corresponding to each subspace decreases, which impairs the ability to effectively model the subspace representations.

\subsection{Ablation of Mask Hyper-parameters}
In this section, we conduct ablation studies to analyze the impact of various hyper-parameters on our method.

\textbf{Mask type.}
We experiment with different masking strategies, including masking only image features, masking only actions, and masking both at random.
As shown on the left side of \cref{fig:ablation_of_mask_type_ratio}, masking only image features or only actions is less effective than masking both simultaneously.
Intuitively, masking both simultaneously leverages the supervisory signals from both modalities, making fuller use of the information contained within the data.
Thus, we simultaneously apply random masking to two modalities.

\begin{figure*}[t!]
\centering
\includegraphics[width=1.5\columnwidth]{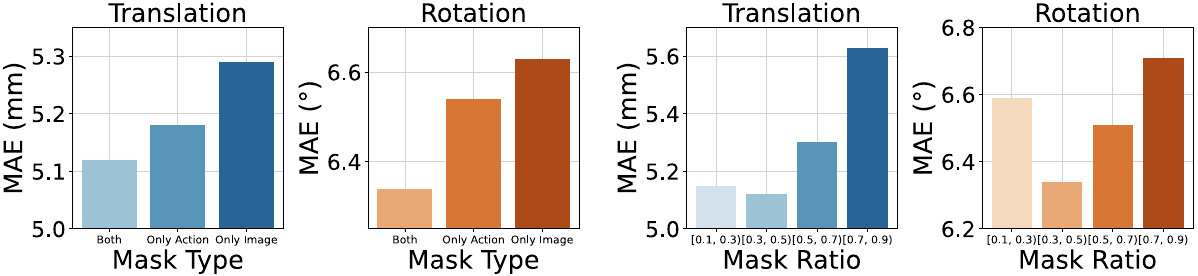}
\caption{
Ablations of masking type and ratio.
}
\label{fig:ablation_of_mask_type_ratio}
\end{figure*}

\textbf{Mask ratio.}
We divided the mask ratio into four intervals from small to large and experimented with the impact of these mask ratios on performance.
As shown on the right side of \cref{fig:ablation_of_mask_type_ratio}, as the mask ratio increases, the prediction error initially decreases but then quickly rises. 
This suggests that when the mask ratio is too high, the model fails to learn useful knowledge.
This occurs because when the mask ratio is too high, the model lacks sufficient inference conditions and can only make random predictions.
For example, if an input sequence leaves only $I_{t_1}$ and requires the prediction of all subsequent masked elements, the potential possibilities are infinite, making it impossible to learn generalizable knowledge.
In contrast, if both $I_{t_1}$ and $a_{t_1 \rightarrow t_2}$ are left, the model can rely on this information to predict the features of $I_{t_2}$, as there is a causal relationship. 
If the model can make correct predictions, it indicates that it has learned spatial relationships between planes.

\begin{figure}[!t]
\centering
\includegraphics[width=1\columnwidth]{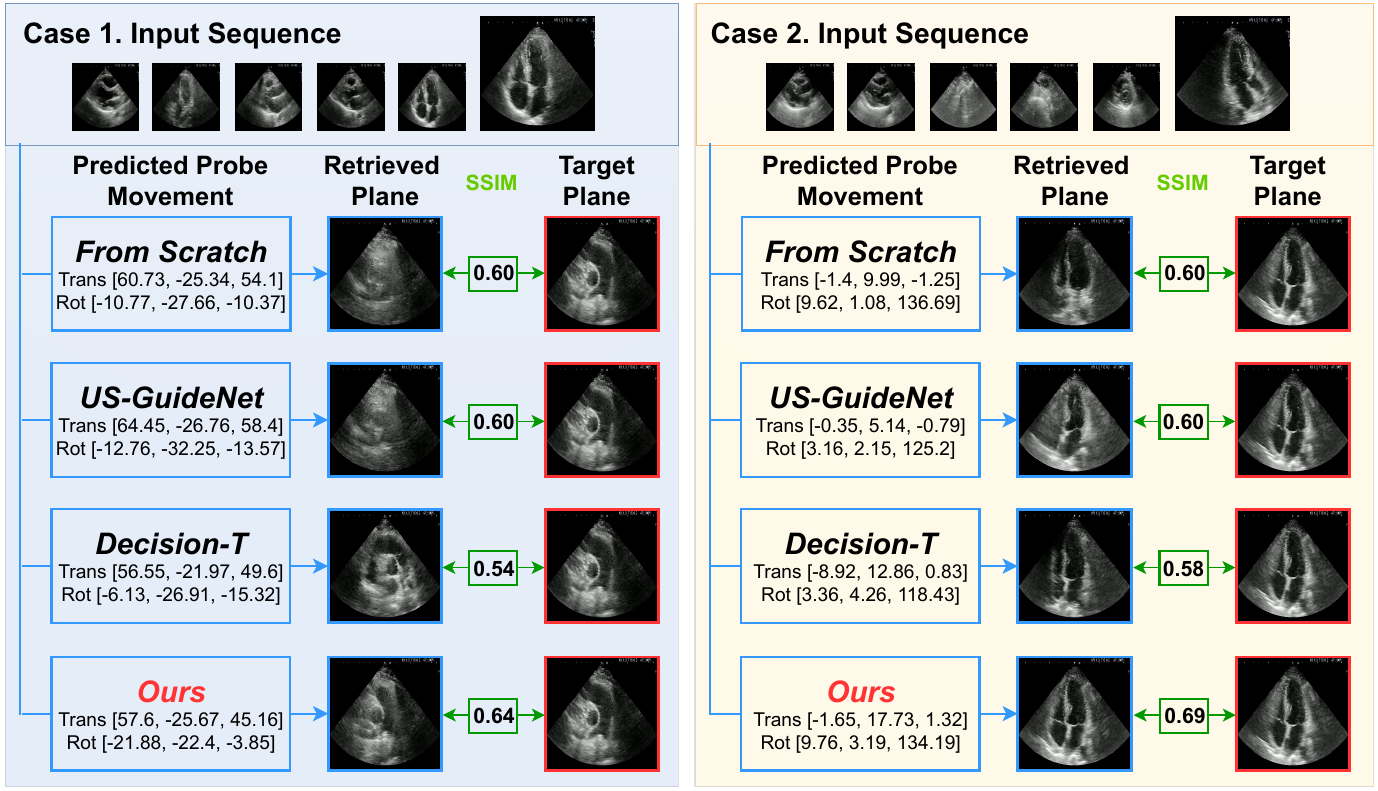}
\caption{
Visualization of the model's prediction.
We performed nearest-neighbor sampling from the scans to retrieve the plane most close to the model's predictions.
The Structural Similarity Index Measure (SSIM) is used to measure the content similarity between the retrieved plane and the target plane.
}
\label{fig:vis}
\end{figure}

\begin{figure}[!t]
\centering
\includegraphics[width=1\columnwidth]{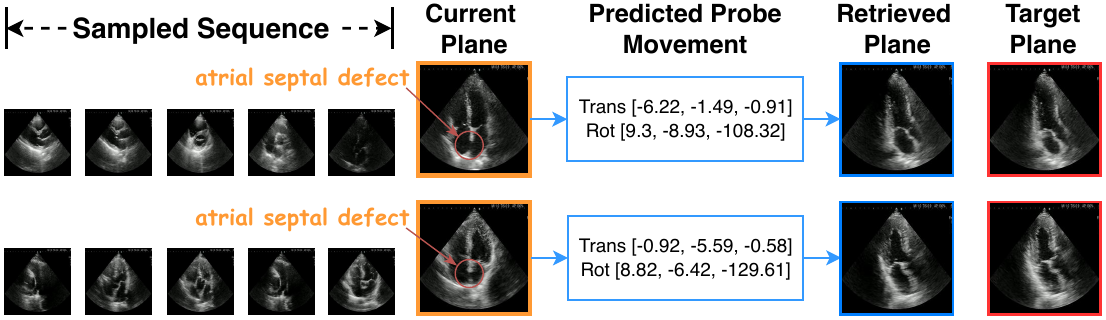}
\caption{
Visualization of the model's prediction on subjects with atrial septal defect.
We performed nearest-neighbor sampling from the scans to retrieve the plane most close to the model's predictions.
}
\label{fig:vis_patient}
\end{figure}

\subsection{Visualization}
To intuitively demonstrate the model's output, we calculate the final plane pose based on the current plane's pose and predicted probe movement action.
Using this pose, we perform nearest-neighbor retrieval from the same scan and present retrieved planes.
We compare four models in \cref{fig:vis}, showing that our model's preditions align closest with the target plane, confirming the superior precision of our guidance signal.
To further validate this, we compare the structural similarity (SSIM) between the retrieved planes and the target plane, demonstrating that our model achieves the highest similarity.

To demonstrate the model's performance on cases with cardiac conditions, we visualized the inference results for two patients with atrial septal defect in \cref{fig:vis_patient}.
The results show a high agreement between the model's output and the expert's annotations, confirming its potential in clinical scenarios involving cardiac diseases.

\subsection{Efficiency}
In ultrasound scanning, highly skilled sonographers make immediate adjustments based on the ultrasound images, a process that involves real-time decision-making. 
Therefore, an ultrasound guidance model must also possess real-time decision-making capabilities to provide timely guidance. 
To this end, we tested the decision-making time for different sequence lengths on a single RTX 3090 and A100 GPU, as shown in \cref{tab:efficiency}.
From the table, it can be seen that our system achieves real-time inference performance on commonly used GPUs.

\begin{table}[!t]
\caption{Efficiency of the guidance model for a single inference.}
\vspace{-10pt}
\begin{center}
\resizebox{1\columnwidth}{!}{
\begin{tabular}{l|p{2cm} p{2cm} p{2cm}} 
    \toprule
Sequence Length & 4 & 6 & 8 \\
    \midrule
RTX3090 & 8.71 ms & 10.56 ms & 12.89 ms \\
A100 & 3.45 ms & 4.94 ms & 6.41 ms \\
    \bottomrule
\end{tabular}
}
\end{center}
\label{tab:efficiency}
\end{table}

\section{Discussion and Conclusion}
In this paper, we propose a cardiac probe movement guidance model which has the potential of driving autonomous robotic echocardiography or helping novice in scanning ten important standard planes.
Specifically, we introduce a novel sequence-aware pre-training method to enhance the model's ability to capture personalized cardiac structures from past sequences, addressing individual variability.
Extensive experiments on a large-scale expert scanning dataset with 1.67M samples demonstrate that our model outperforms other advanced baselines and highlights the superiority of the proposed sequence-aware pre-training method. 

Although experiments have validated the effectiveness of our idea, it should be noted that certain limitations remain and warrant future investigation.
Firstly, the sampling strategy is manually designed, which may be a suboptimal approach.
As evidenced by the ablation study, as the sequence length increases, the rate of error reduction slows down, which may be due to the presence of information redundancy or even noise in the sampled instances.
Therefore, a valuable future direction is to develop methods for sampling the most informative sequences, maximizing information gain from extensive past scanning sequence data.
Secondly, the current work follows an imitation learning strategy, which may be constrained by the operator's expertise. In other words, the learned policy cannot surpass the performance of the sonographer. Given the recent advances in reinforcement learning, it is promising to explore whether such methods could learn an optimal scanning strategy.
Last, information from other modalities such as ultrasound reports and sonographers' gaze information has not yet been collected. 
This type of data has the potential to further enhance the model's performance.
We believe these limitations are also promising research directions that are worth further investigation.

Our work demonstrates the potential of AI-assisted ultrasound scanning, yet several challenges remain for real-world deployment. 
Firstly, integrating such a system into a point-of-care ultrasound platform requires substantially more diverse clinical data to achieve truly robust and clinically viable performance.
Second, beyond assisting human operators, our guidance model can also drive a robotic system. However, developing such a system involves addressing critical human-robot interaction challenges—particularly how to ensure both precise execution of movements and the subject’s comfort and safety.
In conclusion, while this study represents a significant step toward intelligent ultrasound assistance, translating technological innovation into real-world clinical practice demands continued interdisciplinary collaboration—integrating clinical data, AI algorithms, and even robotic control.

\section*{Acknowledgements}
This work is supported by Ministry of Industry and Information Technology of the People's Republic of China (2024YFB4708200).

\bibliographystyle{elsarticle-num} 
\bibliography{reference}

\begin{thebibliography}{10}
\expandafter\ifx\csname url\endcsname\relax
  \def\url#1{\texttt{#1}}\fi
\expandafter\ifx\csname urlprefix\endcsname\relax\def\urlprefix{URL }\fi
\expandafter\ifx\csname href\endcsname\relax
  \def\href#1#2{#2} \def\path#1{#1}\fi

\bibitem{roth2017global}
G.~A. Roth, C.~Johnson, A.~Abajobir, et~al., Global, regional, and national burden of cardiovascular diseases for 10 causes, 1990 to 2015, Journal of the American college of cardiology 70~(1) (2017) 1--25.

\bibitem{mitchell2019guidelines}
C.~Mitchell, P.~S. Rahko, L.~A. Blauwet, et~al., Guidelines for performing a comprehensive transthoracic echocardiographic examination in adults: recommendations from the american society of echocardiography, Journal of the American Society of Echocardiography 32~(1) (2019) 1--64.

\bibitem{braunwald2015war}
E.~Braunwald, The war against heart failure: the lancet lecture, The Lancet 385~(9970) (2015) 812--824.

\bibitem{wild2017large}
P.~S. Wild, J.~F. Felix, A.~Schillert, A.~Teumer, M.-H. Chen, M.~J. Leening, U.~V{\"o}lker, V.~Gro{\ss}mann, J.~A. Brody, M.~R. Irvin, et~al., Large-scale genome-wide analysis identifies genetic variants associated with cardiac structure and function, The Journal of clinical investigation 127~(5) (2017) 1798--1812.

\bibitem{won2024sound}
D.~Won, J.~Walker, R.~Horowitz, S.~Bharadwaj, E.~Carlton, H.~Gabriel, Sound the alarm: The sonographer shortage is echoing across healthcare, Journal of Ultrasound in Medicine (2024).

\bibitem{jiang2024cardiac}
H.~Jiang, Z.~Sun, N.~Jia, M.~Li, Y.~Sun, S.~Luo, S.~Song, G.~Huang, Cardiac copilot: Automatic probe guidance for echocardiography with world model, in: International Conference on Medical Image Computing and Computer-Assisted Intervention, Springer, 2024, pp. 190--199.

\bibitem{narang2021utility}
A.~Narang, R.~Bae, H.~Hong, Y.~Thomas, S.~Surette, C.~Cadieu, A.~Chaudhry, R.~P. Martin, P.~M. McCarthy, D.~S. Rubenson, et~al., Utility of a deep-learning algorithm to guide novices to acquire echocardiograms for limited diagnostic use, JAMA cardiology 6~(6) (2021) 624--632.

\bibitem{bao2024real}
M.~Bao, Y.~Wang, X.~Wei, B.~Jia, X.~Fan, D.~Lu, Y.~Gu, J.~Cheng, Y.~Zhang, C.~Wang, et~al., Real-world visual navigation for cardiac ultrasound view planning, in: International Conference on Medical Image Computing and Computer-Assisted Intervention, Springer, 2024, pp. 317--326.

\bibitem{laumer2023weakly}
F.~Laumer, M.~Amrani, L.~Manduchi, A.~Beuret, L.~Rubi, A.~Dubatovka, C.~M. Matter, J.~M. Buhmann, Weakly supervised inference of personalized heart meshes based on echocardiography videos, Medical image analysis 83 (2023) 102653.

\bibitem{chen20253d}
Z.~Chen, J.~Chen, W.~Zhuo, W.~Xue, D.~Ni, 3d heart reconstruction from sparse pose-agnostic 2d echocardiographic slices (2025).

\bibitem{droste2020automatic}
R.~Droste, L.~Drukker, A.~T. Papageorghiou, J.~A. Noble, Automatic probe movement guidance for freehand obstetric ultrasound, in: International Conference on Medical Image Computing and Computer Assisted Intervention, Springer, 2020, pp. 583--592.

\bibitem{men2023gaze}
Q.~Men, C.~Teng, L.~Drukker, A.~T. Papageorghiou, J.~A. Noble, Gaze-probe joint guidance with multi-task learning in obstetric ultrasound scanning, Medical image analysis 90 (2023) 102981.

\bibitem{assran2023self}
M.~Assran, Q.~Duval, I.~Misra, P.~Bojanowski, P.~Vincent, M.~Rabbat, Y.~LeCun, N.~Ballas, Self-supervised learning from images with a joint-embedding predictive architecture, in: IEEE Conference on Computer Vision and Pattern Recognition, 2023, pp. 15619--15629.

\bibitem{dosovitskiy2020image}
A.~Dosovitskiy, L.~Beyer, A.~Kolesnikov, D.~Weissenborn, X.~Zhai, T.~Unterthiner, M.~Dehghani, M.~Minderer, G.~Heigold, S.~Gelly, J.~Uszkoreit, N.~Houlsby, \href{https://openreview.net/forum?id=YicbFdNTTy}{An image is worth 16x16 words: Transformers for image recognition at scale}, in: International Conference on Learning Representations, 2021.
\newline\urlprefix\url{https://openreview.net/forum?id=YicbFdNTTy}

\bibitem{he2022masked}
K.~He, X.~Chen, S.~Xie, Y.~Li, P.~Doll{\'a}r, R.~Girshick, Masked autoencoders are scalable vision learners, in: IEEE Conference on Computer Vision and Pattern Recognition, 2022, pp. 16000--16009.

\bibitem{huang2017densely}
G.~Huang, Z.~Liu, L.~Van Der~Maaten, K.~Q. Weinberger, Densely connected convolutional networks, in: Proceedings of the IEEE Conference on Computer Vision and Pattern Recognition, 2017, pp. 4700--4708.

\bibitem{ouyang2020video}
D.~Ouyang, B.~He, A.~Ghorbani, N.~Yuan, J.~Ebinger, C.~P. Langlotz, P.~A. Heidenreich, R.~A. Harrington, D.~H. Liang, E.~A. Ashley, et~al., Video-based ai for beat-to-beat assessment of cardiac function, Nature 580~(7802) (2020) 252--256.

\bibitem{zhang2023slide}
W.~Zhang, X.~Ding, Y.~Liu, B.~Qiao, Slide deep reinforcement learning networks: Application for left ventricle segmentation, Pattern Recognition 141 (2023) 109667.

\bibitem{ding2024c2fresmorph}
Y.~Ding, J.~Bu, Z.~Qin, L.~You, M.~Cao, Z.~Qin, M.~Pang, C2fresmorph: A high-performance framework for unsupervised 2d medical image registration, Pattern Recognition 154 (2024) 110615.

\bibitem{amadou2024goal}
A.~A. Amadou, V.~Singh, F.~C. Ghesu, Y.-H. Kim, L.~Stanciulescu, H.~P. Sai, P.~Sharma, A.~Young, R.~Rajani, K.~Rhode, Goal-conditioned reinforcement learning for ultrasound navigation guidance, in: International conference on medical image computing and computer-assisted intervention, Springer, 2024, pp. 319--329.

\bibitem{li2023rl}
K.~Li, A.~Li, Y.~Xu, H.~Xiong, M.~Q.-H. Meng, Rl-tee: Autonomous probe guidance for transesophageal echocardiography based on attention-augmented deep reinforcement learning, IEEE Transactions on Automation Science and Engineering 21~(2) (2023) 1526--1538.

\bibitem{bi2022vesnet}
Y.~Bi, Z.~Jiang, Y.~Gao, T.~Wendler, A.~Karlas, N.~Navab, Vesnet-rl: Simulation-based reinforcement learning for real-world us probe navigation, IEEE Robotics and Automation Letters 7~(3) (2022) 6638--6645.

\bibitem{jiang2025towards}
H.~Jiang, A.~Zhao, Q.~Yang, X.~Yan, T.~Wang, Y.~Wang, N.~Jia, J.~Wang, G.~Wu, Y.~Yue, et~al., Towards expert-level autonomous carotid ultrasonography with large-scale learning-based robotic system, Nature Communications 16~(1) (2025) 7893.

\bibitem{jiang2024structure}
H.~Jiang, M.~Li, Z.~Sun, N.~Jia, Y.~Sun, S.~Luo, S.~Song, G.~Huang, Structure-aware world model for probe guidance via large-scale self-supervised pre-train, in: International Workshop on Advances in Simplifying Medical Ultrasound, Springer, 2024, pp. 58--67.

\bibitem{christensen2024vision}
M.~Christensen, M.~Vukadinovic, N.~Yuan, D.~Ouyang, Vision--language foundation model for echocardiogram interpretation, Nature Medicine (2024) 1--8.

\bibitem{jiao2024usfm}
J.~Jiao, J.~Zhou, X.~Li, M.~Xia, Y.~Huang, L.~Huang, N.~Wang, X.~Zhang, S.~Zhou, Y.~Wang, et~al., Usfm: A universal ultrasound foundation model generalized to tasks and organs towards label efficient image analysis, Medical Image Analysis 96 (2024) 103202.

\bibitem{dong2026hypergraph}
S.~Dong, D.~Li, Y.~Sun, X.~Hu, Y.~Zhao, Hypergraph-driven landmark detection foundation model on echocardiography for cardiac function quantification, Pattern Recognition 170 (2026) 111927.

\bibitem{chen2023rethinking}
G.~Chen, L.~Li, J.~Zhang, Y.~Dai, Rethinking the unpretentious u-net for medical ultrasound image segmentation, Pattern Recognition 142 (2023) 109728.

\bibitem{peng2024multi}
T.~Peng, C.~Wang, C.~Tang, Y.~Gu, J.~Zhao, Q.~Li, J.~Cai, A multi-center study of ultrasound images using a fully automated segmentation architecture, Pattern Recognition 145 (2024) 109925.

\bibitem{wang2016temporal}
L.~Wang, Y.~Xiong, Z.~Wang, Y.~Qiao, D.~Lin, X.~Tang, L.~Van~Gool, Temporal segment networks: Towards good practices for deep action recognition, in: European conference on computer vision, Springer, 2016, pp. 20--36.

\bibitem{chen2021empirical}
X.~Chen, S.~Xie, K.~He, An empirical study of training self-supervised vision transformers, in: IEEE international conference on computer vision, 2021, pp. 9640--9649.

\bibitem{zhang2023biomedclip}
S.~Zhang, Y.~Xu, N.~Usuyama, H.~Xu, J.~Bagga, R.~Tinn, S.~Preston, R.~Rao, M.~Wei, N.~Valluri, et~al., A multimodal biomedical foundation model trained from fifteen million image--text pairs, NEJM AI 2~(1) (2025) AIoa2400640.

\bibitem{mh2024lvm}
D.~MH~Nguyen, H.~Nguyen, N.~Diep, T.~N. Pham, T.~Cao, B.~Nguyen, P.~Swoboda, N.~Ho, S.~Albarqouni, P.~Xie, et~al., Lvm-med: Learning large-scale self-supervised vision models for medical imaging via second-order graph matching, Advances in Neural Information Processing Systems 36 (2024).

\bibitem{touvron2021training}
H.~Touvron, M.~Cord, M.~Douze, F.~Massa, A.~Sablayrolles, H.~J{\'e}gou, Training data-efficient image transformers \& distillation through attention, in: International conference on machine learning, PMLR, 2021, pp. 10347--10357.

\bibitem{oquab2023dinov2}
M.~Oquab, T.~Darcet, T.~Moutakanni, H.~V. Vo, M.~Szafraniec, V.~Khalidov, P.~Fernandez, D.~HAZIZA, F.~Massa, A.~El-Nouby, M.~Assran, N.~Ballas, W.~Galuba, R.~Howes, P.-Y. Huang, S.-W. Li, I.~Misra, M.~Rabbat, V.~Sharma, G.~Synnaeve, H.~Xu, H.~Jegou, J.~Mairal, P.~Labatut, A.~Joulin, P.~Bojanowski, \href{https://openreview.net/forum?id=a68SUt6zFt}{{DINO}v2: Learning robust visual features without supervision}, Transactions on Machine Learning ResearchFeatured Certification (2024).
\newline\urlprefix\url{https://openreview.net/forum?id=a68SUt6zFt}

\bibitem{chen2021decision}
L.~Chen, K.~Lu, A.~Rajeswaran, K.~Lee, A.~Grover, M.~Laskin, P.~Abbeel, A.~Srinivas, I.~Mordatch, Decision transformer: Reinforcement learning via sequence modeling, Advances in neural information processing systems 34 (2021) 15084--15097.

\bibitem{sandler2018mobilenetv2}
M.~Sandler, A.~Howard, M.~Zhu, A.~Zhmoginov, L.-C. Chen, Mobilenetv2: Inverted residuals and linear bottlenecks, in: IEEE Conference on Computer Vision and Pattern Recognition, 2018, pp. 4510--4520.

\bibitem{tran2015learning}
D.~Tran, L.~Bourdev, R.~Fergus, L.~Torresani, M.~Paluri, Learning spatiotemporal features with 3d convolutional networks, in: IEEE international conference on computer vision, 2015, pp. 4489--4497.

\bibitem{vaswani2017attention}
A.~Vaswani, N.~Shazeer, N.~Parmar, J.~Uszkoreit, L.~Jones, A.~N. Gomez, {\L}.~Kaiser, I.~Polosukhin, Attention is all you need, Advances in neural information processing systems 30 (2017).

\end{thebibliography}

\end{document}